%% file: cvpr.tex
\documentclass[final]{cvpr}

\usepackage{times}
\usepackage{epsfig}
\usepackage{graphicx}
\usepackage{amsmath}
\usepackage{amssymb}
\usepackage{booktabs} % for professional tables
\usepackage[pagebackref=true,breaklinks=true,colorlinks,bookmarks=false]{hyperref}

\def\cvprPaperID{7261} % *** Enter the CVPR Paper ID here
\def\confYear{CVPR 2021}

\begin{document}

%%%%%%%%% TITLE
\title{Adversarial Threats to DeepFake Detection: A Practical Perspective}

\author{Paarth Neekhara\\
UC San Diego\\
% {\tt\small pneekhar@ucsd.edu}
\and
Brian Dolhansky\\
Facebook AI\\
\and
Joanna Bitton\\
Facebook AI\\
\and
Cristian Canton Ferrer\\
Facebook AI\\
}

\maketitle

%%%%%%%%% ABSTRACT
\begin{abstract}
% Recent advances in synthetic media techniques have made generation of manipulated content extremely accessible. 
Facially manipulated images and videos or \textit{DeepFakes} can be used maliciously to fuel misinformation or defame individuals. Therefore, detecting DeepFakes is crucial to increase the credibility of social media platforms and other media sharing web sites. State-of-the art DeepFake detection techniques rely on neural network based classification models which are known to be vulnerable to adversarial examples. In this work, we study the vulnerabilities of state-of-the-art DeepFake detection methods from a practical stand point. We perform adversarial attacks on DeepFake detectors in a black box setting where the adversary does not have complete knowledge of the classification models. We study the extent to which adversarial perturbations transfer across different models and propose techniques to improve the transferability of adversarial examples. We also create more accessible attacks using Universal Adversarial Perturbations which pose a very feasible attack scenario since they can be easily shared amongst attackers. We perform our evaluations on the winning entries of the DeepFake Detection Challenge (DFDC) and demonstrate that they can be easily bypassed in a practical attack scenario by designing transferable and accessible adversarial attacks.\footnote{Video Examples: \url{https://deepfakeattacks.github.io/}}
\end{abstract}
\vspace{-2mm}
%%%%%%%%% BODY TEXT
\input{intro}

\section{Background}
\subsection{DeepFakes and DFDC}

\par DeepFakes are a genre of synthetic videos in which a subject's face is swapped with a target face to simulate the target subject in a certain scenario and create convincing footage of events that never occurred~\cite{faceforensicsiccv,advdeepfakes}. Recent video manipulation methods operate end-to-end on a source video and target face and require minimal human expertise to generate fake videos in real-time~\cite{karras2019style,zakharov2019few,nirkin2019fsgan}. 
For expediting research on DeepFake detection, there has been effort in curating datasets~\cite{dolhansky2020deepfake,faceforensicsiccv} of real and fake videos using DeepFake synthesis techniques. 

\par To the best of our knowledge, the recently developed DeepFake Detection Challenge (DFDC) dataset~\cite{dolhansky2019deepfake,dolhansky2020deepfake} is the largest collection of such
real and fake videos, consisting of over 1 million training clips of face swaps produced with a variety of methods. 
For synthesizing the fake videos in the DFDC dataset, 8 different video manipulation techniques were used, many of which are CNN-based techniques. These methods include the traditional DeepFake auto-encoder architecture, a non-learned morphable mask face swap algorithm, and several Generative Adversarial Networks (GAN) techniques like Neural Talking Heads~\cite{zakharov2019few}, FSGAN~\cite{nirkin2019fsgan} and StyleGAN~\cite{karras2019style}. In conjunction with the dataset, a corresponding competition\footnote{https://www.kaggle.com/c/deepfake-detection-challenge} was launched in which competitors were encouraged to submit models trained for DeepFake detection on the training set. These models were then ranked on a hidden, held-out test set, and the winning competitors released their architectures and training strategies publicly.
% Most of these methods are neural network based image

% The public dataset of DFDC was released as a kaggle competetion 
% was organized in which participants trained DeepFake detection methods on curated dataset or Real and Fake videos. Various neural network based facial manipulation techniques like XXXX were used to manipulate videos of . To our knowledge, this is the largest scale study 

% In our work, we generate adversarial examples for fake videos from the DFDC dataset:

\subsection{DeepFake Detection}
Recent state-of-the-art methods for detecting manipulated facial content in videos rely on Convolutional Neural Networks~\cite{afchar2018mesonet,  Amerini_2019_ICCV,li2019exposing,rahmouni2017distinguishing,faceforensicsiccv} to distinguish AI-generated fake videos from real videos. These methods model the DeepFake identification problem as a per-frame classification problem. They employ typical image classification networks that either operate on the entire frame or on a cropped portion of the frame that has domain specific information. For example, state-of-the-art DeepFake classification systems~\cite{nlab, wm,selim} consist of a face-tracking method, following by the cropped face passed on to a CNN-based classifier for classification as Real or Fake~\cite{afchar2018mesonet,chollet2017xception}. The final label of a video is usually the aggregation of the labels for some candidate frames of the video.

While it seems intuitive that exploiting temporal dependencies using sequence models should improve a detector's ability to spot manipulated videos, 
the insights from the results of the DFDC challenge~\cite{dolhansky2020deepfake,dolhansky2019deepfake} show that the best performing models operate on a frame level. In fact, the winning team~\cite{selim} of the DFDC challenge explicitly noted that ideas besides the frame-by-frame detector did not improve their performance on the public leader-board. 
One reason we anticipate for this is that the recent DeepFake generation techniques have improved temporal consistencies in the videos; however creating plausible face-swaps in images is still challenging due to the artifacts introduced by upsampling methods in autoencoders.

% The third place winner~\cite{nlab} of the DFDC also used a 3D CNN based sequence model, however, the gains 

\begin{figure}[t]
    \centering
    \includegraphics[width=1.0\columnwidth]{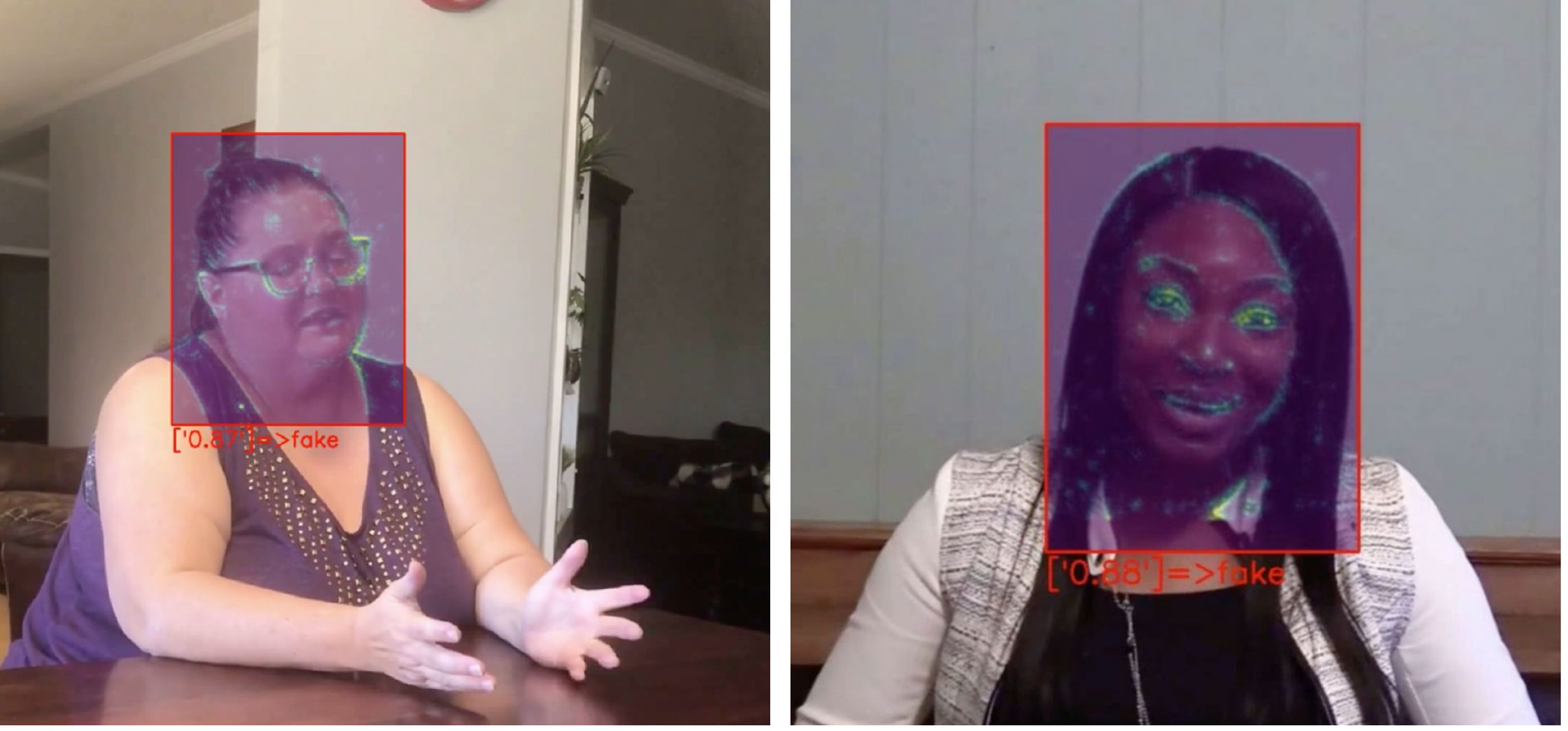}{\centering}
    \caption{Gradient saliency maps obtained on DeepFake videos using guideded-backpropogation on a CNN-based detector~\cite{selim}. The highlighted areas indicate the image regions that strongly influence the detector's predictions.}
    \label{fig:explain}
\end{figure}

In our work, we focus on the top three winning entries~\cite{nlab, wm,selim} of the DFDC challenge. In order to understand their vulnerabilities, we first studied the commonalities across these detection methods. 
Our objective was to gain insight into what the detector is looking at when it makes a decision about a video being Real or Fake. This is typically done by obtaining the gradient of the score of the predicted class with respect to the input image and plotting the magnitude of these gradients as a heat-map. 
Back-propagating gradients naively does not result in  interpretable visualizations. This is because we only care about pixels which activate a neuron rather than suppress it (suppression is indicated by negative gradients)~\cite{gbb}. 
Therefore, we use guided backpropagation which defines custom gradient estimates for activation functions like ReLU and suppresses negative gradients during the backward pass~\cite{gbb}. 
We then standardize the gradient obtained with respect to the input and overlay the heat-map on the frame to visualize the areas of the image that trigger the network’s output. Figure~\ref{fig:explain} shows some examples of the saliency maps obtained while analyzing two different detectors on DeepFake videos.

Our initial observations on these saliency maps suggested that different CNN-based detection methods attend to similar aspects of the input frame for predicting the label. 
These aspects include the edges of the face, the eyes, lips, teeth etc. 
These similarities across different detection methods indicate that adversarially modifying such aspects of the image could potentially fool multiple detection methods. 
We validate this hypothesis in our work by studying the transferability of adversarial examples (Section~\ref{sec:exps}) across different detection methods and proposing techniques (Section~\ref{sec:transferattacks}) that improve the transferability.
% In our experiments, we empirically find that this is inface

\subsection{Prior Work on Fooling DeepFake Detectors}
\label{sec:pastfooling}
Neural network-based classification systems have been known to be vulnerable to adversarial examples. Adversarial examples are intentionally designed inputs to a machine learning (ML) model that cause the model to make a mistake~\cite{42503}. Prior work has shown that gradient-based attacks can effectively fool neural networks~\cite{carlini2017towards,goodfellow2014explaining,universal,papernot1,limitations,shi2019curls,tramer} with a minimal amount of perturbation added to the original input. 

Recently, gradient-based adversarial attacks have also been applied on CNN-based DeepFake detection systems to expose their vulnerabilities to adversarial examples~\cite{carlini2020evading,gandhi2020adversarial,advdeepfakes}. While some of this past work~\cite{carlini2020evading,gandhi2020adversarial} focuses on attacks on image classification models, the authors of~\cite{advdeepfakes} study the vulnerability of video DeepFake detection methods which follow the same detection pipeline as the methods studied in our work. While this past work demonstrates that adversarial examples can fool video DeepFake detectors, designing such adversarial videos requires complete access to the victim model architecture and parameters (white-box attack). This assumption makes the threat very limited in a real-world scenario since the model architecture and parameters can be kept hidden by the service provider. The black-box attacks proposed in~\cite{advdeepfakes} require querying the victim model multiple times and access to the raw scores given by the classifier for each frame the attacker wishes to misclassify. 
The number of queries and access to raw classifier scores can be restricted to thwart the adversary thereby preventing this black-box attack.

Adversarial examples pose a practical threat to DeepFake detection if they are transferable across different detection methods. That is, if adversarial videos designed to fool some open source DeepFake detection method can also reliably fool other unseen CNN-based detection methods, it poses a real security threat to deploying CNN-based detectors in production.  Several past works have studied this \textit{transferability} property of adversarial examples 
% against neural network classifiers - transferable black-box attacks, 
where an attacker first generates an adversarial perturbation on a (white-box) surrogate source model, and then transfers it to the unknown target network~\cite{cheng2019improving,goodfellow2014explaining,transfer2, papernot2016transferability,papernot2017practical,szegedy2013intriguing,zhou2018transferable}. 

In our work, we study the transferability of adversarial examples across various DeepFake detection methods. We find that differences in 1) input-preprocessing steps and 2) face detection methods across DeepFake detectors hamper the cross-model transferability of adversarial examples.
% make the goal of transferability of adversarial examples more challenging in the DeepFake detection domain. 
We propose techniques to overcome these challenges in Section~\ref{sec:transferattacks}, and further propose more accessible attacks inspired from universal adversarial perturbations~\cite{behjati2019universal,Liu_2019_ICCV,mopuri2017fast,neekhara2019universal,reddy2018ask}. Universal adversarial perturbations pose a more practical threat to DeepFake detection since they can be easily shared amongst attackers and require no technical expertise in adversarial machine learning.

\section{Methodology}

\subsection{Threat Model}
Given a video (\textit{Real} or \textit{Fake}), our task is to adversarially modify the video such that the label predicted by a victim DeepFake detection method is incorrect. That is, we want to modify the videos such that the \textit{Fake} videos are classified as \textit{Real} and vice-versa. Misclassifying a \textit{Fake} video as \textit{Real} can be used by the adversary to propagate false information. 
Misclassifying a \textit{Real} video as \textit{Fake} can be used by the adversary to cover up an event that did actually happen. 

\begin{figure}[!t]
    \centering
    \includegraphics[width=1.0\columnwidth]{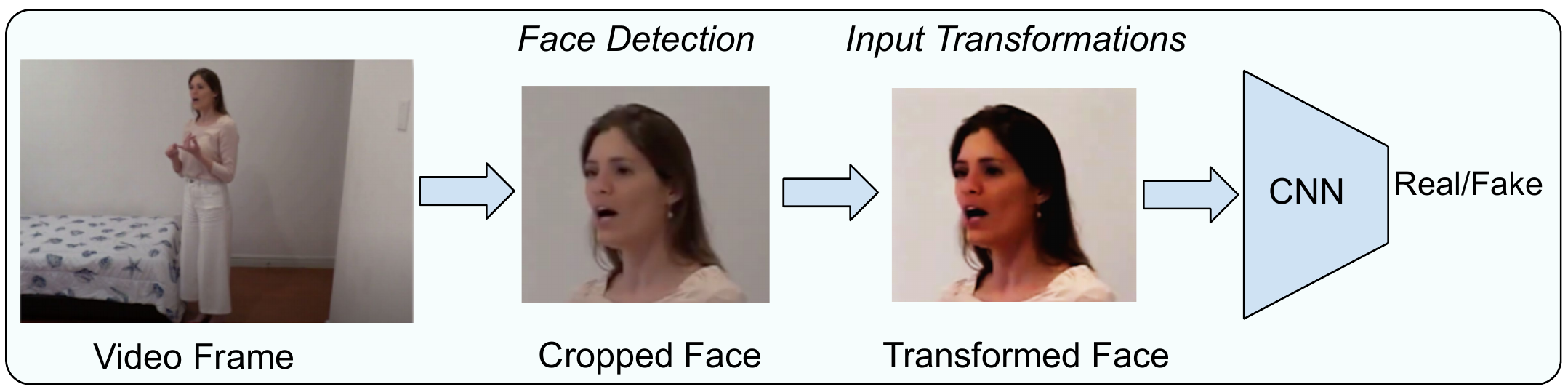}{\centering}
    \caption{A typical DeepFake detection pipeline. A face-tracking model crops the face from all video frames, which is resized and normalized appropriately to be fed as input to a CNN classification model.}
    \label{fig:dfdpipeline}
\end{figure}

The goal is to also ensure imperceptibility of the adversarial perturbation. In the image domain, $L_p$ norms are commonly used to quantify the amount of perturbation added to create an adversarial input. The authors of~\cite{goodfellow2014explaining} recommend constraining the maximum distortion of any individual pixel in the image using the $L_\infty$ metric. In the attacks discussed in this paper, the adversarial perturbation is added to each frame of the input video to create the adversarially modified video. Following past work on fooling DeepFake detectors ~\cite{advdeepfakes}, we use the $L_\infty$ metric to constrain the the amount of distortion added to each frame.

\noindent \textbf{Notation:} We follow the notation previously used in~\cite{carlini2017towards,papernot2016distillation,advdeepfakes}; we define $F$ to be the full neural network (classifier) including the softmax function, $Z(x) = z$ to be the output of all layers except the softmax (that is $z$ are the logits). That is: 
$$F(x) = softmax(Z(x)) = y$$
The classifier assigns the label $C(x) = \arg\max_i(F(x)_i)$ to input frame $x$.

\noindent \textbf{Problem Formulation:} Mathematically, for each video frame $x$, we aim to find an adversarial frame $x_{adv}$ such that:
$$C(x_{\mathit{adv}}) = \mathit{y} \text{ and } ||x_{\mathit{adv}}-x_0||_\infty < \epsilon$$ 

where $y$ is the target label. In our case the target label is \textit{Real} for \textit{Fake} videos and \textit{Fake} for \textit{Real} videos. 
In the upcoming sections, we study this attack goal in various attacker knowledge settings and constraints. 

\subsection{Victim Models}
\label{sec:victimmodels}
The victim models we consider in our work model DeepFake detection as a per-frame classification problem. These models further decompose the frame classification problem into the following two steps:
\begin{enumerate}
    \item A face tracking model detects the bounding box of the face in each frame.
    \item The cropped face is then pre-processed using some input transformations (e.g. resizing, center-cropping and normalization) and fed as input to a CNN classification model that scores the frame as \textit{Real} or \textit{Fake.}
\end{enumerate}

Finally, the scores of all or a subset of the frames are aggregated to obtain the final label of the video. The above detection pipeline has been used by the top 5 winning entries of the DFDC challenge. The CNN architectures and data-augmentation procedures vary across different methods. Table~\ref{tab:VictimModels} details various DeepFake detection methods considered in our work with their respective face detection methods and CNN architectures used for classification.

\subsection{White-box attacks}
\label{sec:whitebox}
In this setting, we assume that the attacker has complete access to the detection model, including the face extraction pipeline and the architecture and parameters of the classification model. To construct adversarial examples using the attack pipeline described above, we use the iterative gradient sign method~\cite{kurakin2016adversarial} to optimize the following objective:
\begin{equation}
\begin{split}
& \text{Minimize } \mathit{loss}(x') \text{ where}\\
& \mathit{loss}(x') = \mathit{max}(Z(x')_{\mathit{o}} - Z(x')_{\mathit{y}}, 0) \\
\end{split}
\label{eq1}
\end{equation}

Here, $Z(x)_y$ is the final score for target label $y$ and $Z(x)_o$ is the score of the original label $o$ before the softmax operation in the classifier $C$. The loss function we use is recommended 
%by Carlini \emph{et al.} 
by~\cite{carlini2017towards} because it is empirically found to generate less distorted adversarial samples and is robust against defensive distillation. 
We use the iterative gradient sign method to optimize the above loss function while constraining the magnitude of the perturbation as follows: 
\begin{equation}
\begin{split}
% & x'_{0} = x_{0} \\
& x_i = x_{i-1} - \text{clip}_{\epsilon}(\alpha \cdot \text{sign}(\nabla \mathit{loss}(x_{i-1})))\\
\end{split}
\label{eq2}
\end{equation}
We continue gradient descent iterations  until success or until a given number number of maximum iterations, whichever occurs earlier. 
We solve the optimization problem for each frame of the given video and combine all the adversarial frames together to generate the adversarial video. 
In our experiments, we demonstrate that we are able to successfully fool all the detection methods studied in our work in the white-box attack setting using the above attack. 
However, the transferability of adversarial examples generated using this attack across different methods is limited. In the next section we propose techniques to overcome this challenge. 

% In our experiments, we demonstrate that while we are able to achieve an average attack success rate of 99.05\% when we save videos with uncompressed frames, the perturbation is not robust against video compression codecs like \textit{MJPEG}.  
% The success rate drops significantly when we apply such commonly used video compression codecs while saving adversarial videos. 
% In the following section, we discuss our approach to overcome this limitation of our attack. 

\subsection{Black-box: Transfer attacks}
\label{sec:transferattacks}
Past works (Section~\ref{sec:pastfooling}) have studied that adversarial inputs can transfer across different models. That is, an adversarial input that was designed to fool a particular victim model can possibly fool other models that were trained for the same task. This is because different models learn similar decision boundaries and therefore have similar vulnerabilities.
However, for DeepFake detectors, the goal of making transferable adversarial videos is more challenging due to multiple steps involved in the DeepFake detection pipeline and the differences in these steps across various methods.
\begin{itemize}
    \item Different face detection methods result in different face-crops. 
    \item Different data-augmentation procedures during training result in different levels of robustness to adversarial examples.
    \item Different input pre-processing pipelines, such as image resizing, cropping and normalization parameters, vary across different detection methods. 
\end{itemize}

Therefore, to craft \textit{transferable} adversarial videos, it is important to ensure robustness to such differences across various methods. To accomplish this, we craft adversarial examples that are robust over a given distribution of input transformations~\cite{eot}. Given a distribution of input transformations $T$, input image $x$, and target class $y$, our objective is as follows:
$$ x_{adv} = \mathit{argmax}_{x} \mathbb{E}_{t\sim T} [F(t(x))_y] \text{ s.t. } ||x - x_0||_\infty < \epsilon $$
That is, we want to maximize the expected probability of target class $y$ over the distribution of input transforms $T$. To solve the above problem, we update the loss function given in Equation~\ref{eq1} to be an expectation over input transforms $T$ as follows:
$$\mathit{loss}(x) = \mathbb{E}_{t\sim T} [\mathit{max}(Z(t(x))_{\mathit{o}} - Z(t(x))_{\mathit{y}}, 0)]$$
Following the law of large numbers, we estimate the above loss functions for $n$ samples as:
\begin{equation}
\mathit{loss}(x) = \frac{1}{n}\sum_{t_i\sim T} [\mathit{max}(Z(t_i(x))_{\mathit{o}} - Z(t_i(x))_{\mathit{y}}, 0)]
\label{eq:expectation}
\end{equation}

Since the above loss function is a sum of differentiable functions, it is tractable to compute the gradient of the loss w.r.t.~to the input $x$. We minimize this loss using the iterative gradient sign method given by Equation~\ref{eq2}. We iterate until a maximum number of iterations is reached or until the attack is successful under the sampled set of transformation functions, whichever happens first.

Next, we describe the class of input transformation functions we consider for the distribution $T$:

\begin{itemize}
    \item Translation: We pad the image on all four sides by zeros and shift the pixels horizontally and vertically by a given amount. This transform ensures robustness to different face-detection and cropping pipelines across various methods.
Let $t_x$ be the transform in the $x$ axis and $t_y$ be the transform in the $y$ axis, then $t(x) = x'_{H,W,C}$ s.t $x'[i,j,c] = x[i+t_x,j+t_y,c]$
    
    \item Downsizing and Upsizing: The image is first downsized by a factor $r$ and then up-sampled by the same factor using bilinear re-sampling.
    
    \item Gaussian Noise Addition: Addition of Gaussian noise sampled from $\Theta\sim{\mathcal{N}(0,\sigma)}$ to the input image. This transform is given by $t(x)= x + \Theta $ 
    
\end{itemize}

% \textbf{Gaussian Blur:}  Convolution of the original image with a Gaussian kernel $\mathit{k}$. This transform is given by $t(x)= k \ast x $ where $\ast$ is the convolution operator.

The details of the hyper-parameter search distribution used for these transforms can be found in Section~\ref{sec:exptransfer}. Empirically, we find that ensuring robustness of adversarial examples significantly improves attack transferability across various detection methods.

\subsection{Universal attacks}
\label{sec:universalmethod}

While the transferability of adversarial perturbations poses a practical threat to DeepFake detectors in production, creating an adversarial video requires significant technical expertise in adversarial machine learning - the attacker needs to solve an optimization problem for each frame of the video to fool the detector. 

To ease the process of fooling DeepFake detectors, we aim to design more accessible adversarial attacks that can be easily shared amongst attackers. Past works~\cite{universal,behjati2019universal,neekhara2019universal} have shown the existence of universal adversarial perturbations that can fool classification models in various input domains. 
We aim to find a single universal adversarial perturbation which when added across all frames of any video, will cause the victim DeepFake Detector to classify the video to a target label. 

That is, we aim to find a targeted universal perturbation $\delta$ such that:
\begin{equation}
\begin{split}
& C(x + \delta) = y \quad \textit{s.t} \quad ||\delta||_\infty < \epsilon\\
& \text{for ``most'' $x$ in our dataset}
% & | \\
\end{split}
\label{universal}
\end{equation}

where $y$ is the target class. We train separate perturbations for Real and Fake target labels.
In order to ensure robustness to differences across detection methods, we incorporate  the transformation functions described in Section ~\ref{sec:transferattacks}. We train the universal adversarial perturbation on a dataset of videos that are labelled opposite from our target label. On this dataset of videos, we aim to maximize the log-likelihood of predicting our target label $y$. 
Additionally to ensure the imperceptibility of the adversarial perturbation we penalize the $L_2$ distortion of the perturbation by adding a regularization term in our objective.
Thus, our final objective to train the a universal perturbation for a target label $y$ is as follows:
\begin{equation}
\begin{split}
& \textit{Minimize} \sum_{x \text{ in } D} \mathbb{E}_{t\sim T} [L(F(t(x + \delta)), y)] + c||\delta||_2 \\
& \quad \quad \quad  \quad \quad \quad \textit{ such that } \quad ||\delta||_\infty < \epsilon\\
% & | \\
\end{split}
\label{eq:universalobjective}
\end{equation}
% \textbf{Universal Adversarial Perturbations}\\
% \textbf{Adversarial Transformation Networks}

Here, $L$ is the cross-entropy loss between the predictions and our target label, $c$ is a hyper-parameter to control the regularization loss and $x$ is an input frame of a video from our dataset $D$. 
% After each iteration, we clip the perturbation to ensure that that $||\delta||_\infty < \epsilon$. 
Similar to Equation~\ref{eq:expectation}, we estimate the above expectation using $n$ samples as follows:
\begin{equation}
\mathbb{E}_{t\sim T} [L(F(t(x + \delta)), y)] = \frac{1}{n}\sum_{t_i\sim T} [L(F(t_i(x + \delta)), y)]
\label{eq:expectation2}
\end{equation}

To ensure the constrain $||\delta||_\infty < \epsilon$, we express $\delta$ as follows:
$$\delta = \epsilon\cdot\mathit{tanh}(p)$$ where $p$ is a trainable unconstrained parameter having the same dimensions as $\delta$. 
We fix the size of the perturbation vector $p$ to be $3\times256\times256$ in our experiments, but resize the perturbation using bilinear interpolation to match the size of our input $x$. We iteratively optimize the objective given by Equation~\ref{eq:universalobjective} using gradient descent. In our experiments, we find that targeting certain DeepFake detectors not only results in input-agnostic universal perturbations but also model-agnostic universal perturbations.

\section{Experiments}
\label{sec:exps}
\subsection{Experimental Setup}
\noindent \textbf{DeepFake detectors:} In our work, we consider the DeepFake detection methods proposed by the top three winning entries of DFDC~\cite{dolhansky2019deepfake}. All of these methods follow the DeepFake detection pipeline described in Section~\ref{sec:victimmodels}. However, these methods use different CNN model architectures for classification and face-tracking. 
Table~\ref{tab:VictimModels} lists the DeepFake detection methods studied in this work along with their respective CNN architectures used for classification and face detection. In our experiments we use the terms \textit{victim model} and \textit{test model} and define them as:

\begin{itemize}
    \item \textit{Victim model:} The detection model that the attack/adversarial perturbation is trained on, in the complete-knowledge (white-box) attack scenario.
    \item \textit{Test model:} The model on which we evaluate the attack - Can be the same as the victim model (white-box) or an unseen detection model (black-box).
\end{itemize}

\setlength\tabcolsep{6pt} % default value: 6pt
\begin{table}[!t]
\centering
\resizebox{\columnwidth}{!}{%
\begin{tabular}{@{}l|ccc|c@{}}
\toprule
\textbf{Model} &  \textit{Team Name} & \textit{Classifier} & \textit{Face detection} & \textit{AUC}\\ 
\midrule
EN-B7 Selim~\cite{selim} & Selim & EfficientNet B7~\cite{tan2019efficientnet} & MTCNN~\cite{mtcnn} & 0.717\\
XN WM~\cite{wm} & Team WM & XceptionNet~\cite{chollet2017xception} & RetinaFace~\cite{retinaface} & 0.724\\
EN-B3 WM~\cite{wm} & Team WM & EfficientNet B3~\cite{tan2019efficientnet} & RetinaFace~\cite{retinaface} & 0.724\\
EN-B7 NLab~\cite{nlab} & NTech Lab & EfficientNet B7~\cite{tan2019efficientnet} & DSFD~\cite{dsfd} & 0.717\\
\bottomrule
\multicolumn{1}{c}{}
\end{tabular}%
}
\caption{Different DeepFake detection systems studied in our work with their respective classification models, face detection models and detection AUC scores on the DFDC test set.}
\label{tab:VictimModels}
\end{table}

\noindent \textbf{Datasets:} We craft adversarial videos for the first 100 Fake and 100 Real videos in the public DFDC validation set~\cite{dolhansky2019deepfake}. These videos contain a total of 30,300  frames. The videos are recorded in various lighting and background conditions and include people with different skin-tones. 
\\
\\
\textbf{Evaluation Metrics:}
After performing our attacks we combine the adversarial frames to create the adversarial video.
We report the following metrics for evaluating our
attacks:
\\
\textit{Success Rate (SR)}: The percentage of videos for which we are able to successfully flip the original correct label predicted by a given detection method. Note that we do not take into account the videos that are originally mis-predicted by the classifier.
\\
\textit{Mean distortion ($L_\infty$)}: The average $L_\infty$ distortion between the adversarial and original frames. The pixel values are scaled in the range [0,1], so changing a pixel from full-on to full-off in a grayscale image would result in $L_\infty$ distortion of 1 (not 255). 

\subsection{White-box attacks}
In this section, we discuss the evaluation of the iterative gradient based white-box attack described in Section~\ref{sec:whitebox}. Following past work~\cite{advdeepfakes}, we set the max allowed $L_\infty$ norm $\epsilon$ as $16/255$ and continue the attack iterations until the predicted score of our target label is greater than $0.99$.

As shown in  Table~\ref{tab:whitebox}, for a given victim model, we are able to achieve 100\% success rate for the same test model.
EfficientNet-B7 by NTech Lab requires the highest amount of adversarial perturbation under the $L_\infty$ metric as compared to other methods in this study.
We also evaluate the extent to which these perturbations transfer across different methods. We find that perturbations trained to fool EfficientNet-B7 by Team NTech Lab result in the most transferable attacks as indicated by the higher success rates on other test models. This suggests that \textit{EN-B7 NLab} is relatively more robust to adversarial perturbations in comparison to the other models used in this study (also indicated by higher $L_\infty$ perturbation required to fool \textit{EN-B7 NLab}). 
% This is possibly due to more number of image augmentations used during training of the detector~\cite{nlab}.

% Figure XX shows an example of original and adversarial frames

\begin{figure}[!t]
    \centering
    \includegraphics[width=1.0\columnwidth]{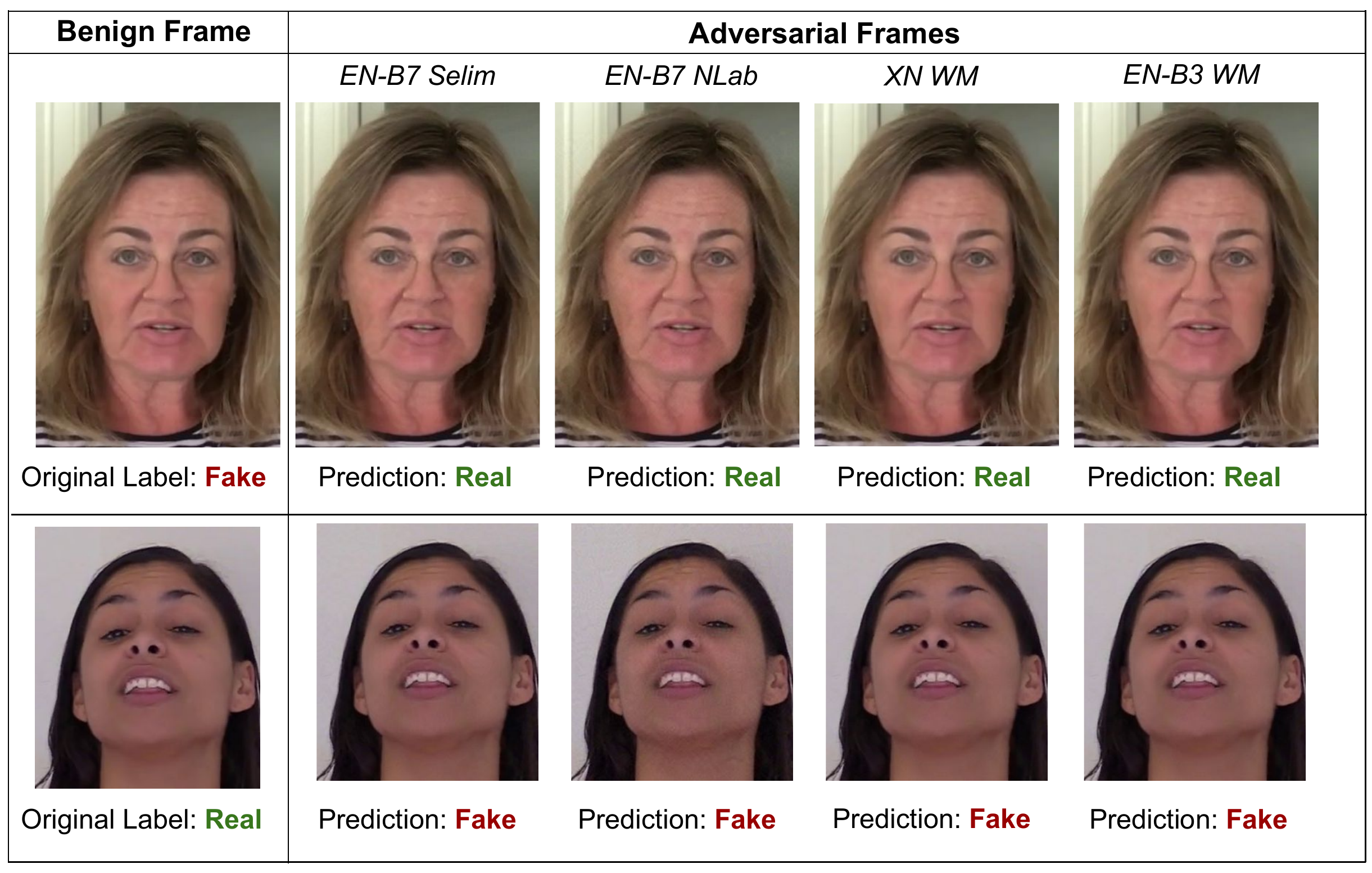}{\centering}
    \caption{Examples of benign and adversarial video frames generated using our simple white-box attack targeting different DeepFake detection methods. It can be seen that the original prediction and label of the video can be successfully flipped by adding an imperceptible amount of adversarial perturbation in a white-box attack setting. }
    \label{fig:whiteboxexamples}
\end{figure}

Figure~\ref{fig:whiteboxexamples} shows examples of adversarial faces generated to flip the original label of a given frame while targeting different victim models. As indicated by the low $L_\infty$ norm of the perturbation, the amount of added perturbation is very imperceptible indicating that DeepFake detectors are extremely vulnerable to adversarial examples and can be easily fooled. Video examples are linked in the footnote on the first page.

\subsection{Transfer attacks}
\label{sec:exptransfer}
To improve the transferability of adversarial examples across different methods, we perform our transfer attack described in Section~\ref{sec:transferattacks} and evaluate the adversarial videos against unseen detection methods in a black-box setting. The hyper-parameters of the transformation functions used for the attack have been provided in Table~\ref{tab:transforms}. All other attack hyper-parameters are kept the same as our simple white-box attack.

As indicated by the results in Table~\ref{tab:transferattacks}, we are able to significantly improve the transferability of adversarial perturbations across different detection methods as compared to our simple white-box attack. 
% Specifically, while attacking the most robust \textit{EN-B7 NLab} model . 
The adversarial perturbations are most transferable across models with the same architecture. For example, we are able to achieve high cross-transferability between \textit{EN-B7 Selim} vs \textit{EN-B7 NLab}. Similar to our observation in the previous section, attacking \textit{EN-B7 NLab} results in the most transferable adversarial attacks - we are able to achieve at least 72\% success rate across all other detection methods when attacking \textit{EN-B7 NLab}.

\begin{table}[!t]
\centering
\resizebox{\columnwidth}{!}{%
\begin{tabular}{@{}l|c@{}}
\toprule
\textbf{Transform} &  \text{Hyper-parameter search distribution}\\ 
\midrule
\textit{Translation} & $d_x \sim \mathcal{U}[-20, 20]$, $d_y \sim \mathcal{U}[-20, 20]$\\
\textit{Gaussian Noise} & $\sigma \sim \mathcal{U}[0.05, 0.07] $\\
\textit{Down-sizing \& Up-sizing} & Scaling factor $r \sim \mathcal{U}[2, 5] $\\
\bottomrule
\multicolumn{1}{c}{}
\end{tabular}%
}
\caption{Search distribution of hyper-parameters of different transformations used for our transfer attack. During training, we sample three functions from each of the transforms to estimate the gradient of our expectation over transforms.}
\label{tab:transforms}
\end{table}

\begin{figure}[!t]
    \centering
    \includegraphics[width=1.0\columnwidth]{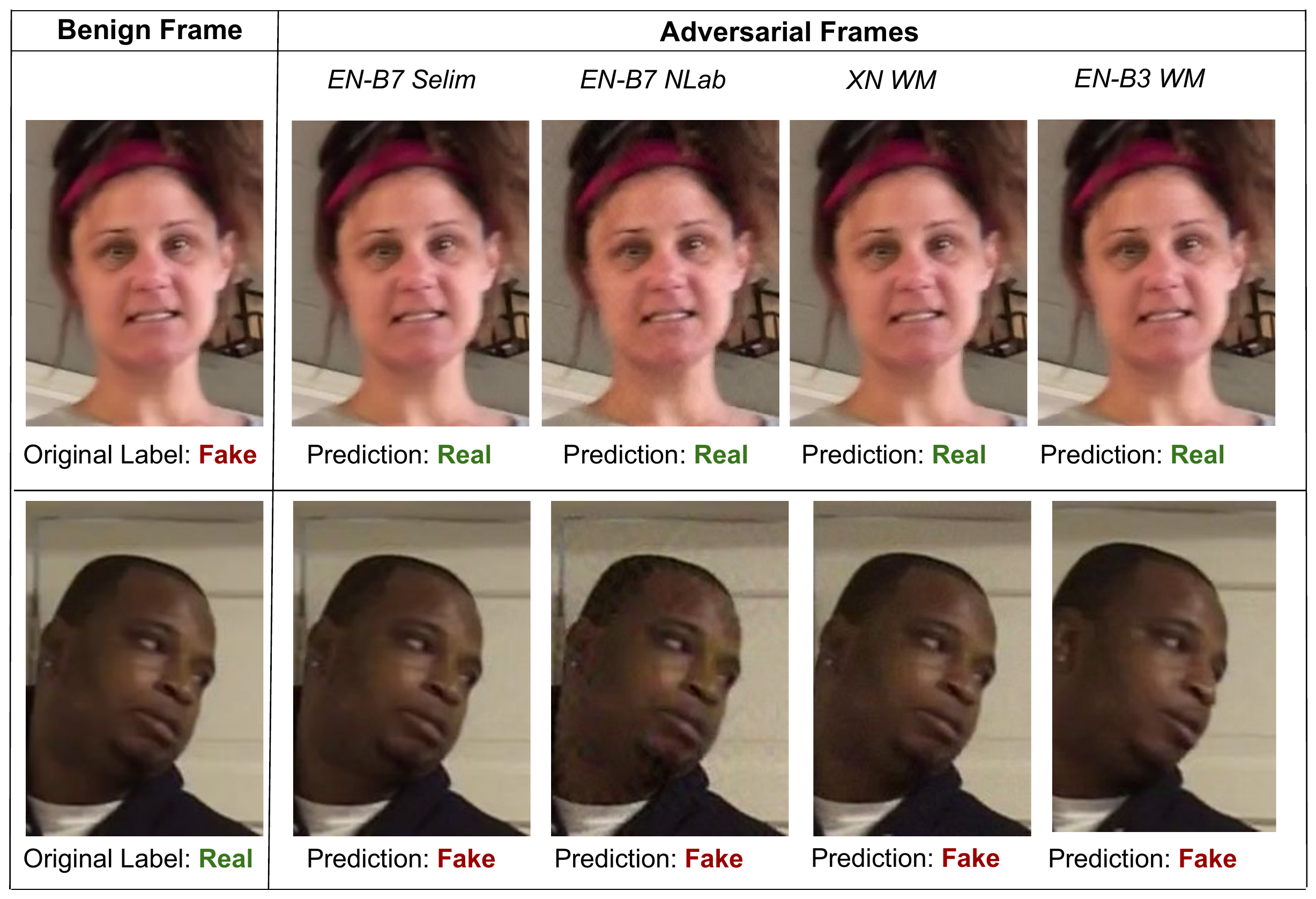}{\centering}
    \caption{Examples of benign and adversarial video frames generated using our transfer attack targeting different DeepFake detection methods.
    }
    \label{fig:transferexamples}
\end{figure}

In order to ensure the robustness of the adversarial perturbation to input transformations, a relatively higher amount of perturbation is required as compared to our white-box attack (Mean $L_\infty$ distortion 0.0135 vs 0.0077).   Figure~\ref{fig:transferexamples} shows examples of adversarial faces generated to flip the original label of a given frame while targeting different victim models using our transfer attack. While a higher amount of perturbation is required as compared to our simple white-box attack, visually the  perturbation is still fairly imperceptible.

\setlength\tabcolsep{2pt} % default value: 6pt
\begin{table}[h]
\centering
\resizebox{\columnwidth}{!}{%
\begin{tabular}{@{}l|c|c|c|c|c@{}}
\multicolumn{2}{c}{}& \multicolumn{4}{c}{\textit{Test Models}}\\ 
\toprule
\textbf{Victim Model} &  \textbf{$L_\infty$} & EN-B7 Selim & EN-B7 NLab & XN WM & EN-B3 WM\\ 
\midrule
EN-B7 Selim & 0.007 & 100.0\% & 59.5\% & 57.0\% & 38.5\%\\
EN-B7 NLab & 0.013 & 94.0\% & 100.0\% & 66.5\% & 49.5\%\\
XN WM & 0.006 & 13.0\% & 12.5\% & 100.0\% & 12.0\%\\
EN-B3 WM & 0.005 & 21.0\% & 15.5\% & 22.0\% & 100.0\%\\
\bottomrule
\multicolumn{1}{c}{}
\end{tabular}%
}
\caption{Attack success rates of the white-box attacks (Section~\ref{sec:whitebox}) on different victim models and their transferability to unseen detectors (test models). 
% The highlighted entry in the table shows the attack success rate of targeting \textit{} and testing on \textit{} 
}
\label{tab:whitebox}
% \vspace{-4mm}
\end{table}

\setlength\tabcolsep{2pt} % default value: 6pt
\begin{table}[h]
\centering
\resizebox{\columnwidth}{!}{%
\begin{tabular}{@{}c|c|c|c|c|c@{}}
\multicolumn{2}{c}{}& \multicolumn{4}{c}{\textit{Test Models}}\\ 
\toprule
\textbf{Victim Model} &  \textbf{$L_\infty$} & EN-B7 Selim & EN-B7 NLab & XN WM & EN-B3 WM\\ 
\midrule
EN-B7 Selim & 0.010 & 100.0\% & 89.0\% & 72.5\% & 62.0\%\\
EN-B7 NLab & 0.018 & 99.0\% & 100.0\% & 72.0\% & 76.5\%\\
XN WM & 0.018 & 49.0\% & 33.5\% & 100.0\% & 46.0\%\\
EN-B3 WM & 0.008 & 46.5\% & 35.0\% & 47.5\% & 100.0\%\\
\bottomrule
\multicolumn{1}{c}{}
\end{tabular}%
}
\caption{Attack success rates of the transfer attacks (Section~\ref{sec:transferattacks}) on different victim models and their transferability to unseen detectors (test models).}
\label{tab:transferattacks}
\end{table}

\setlength\tabcolsep{2pt} % default value: 6pt
\begin{table}[h]
\centering
\resizebox{\columnwidth}{!}{%
\begin{tabular}{@{}c|c|c|c|c|c@{}}
\multicolumn{2}{c}{}& \multicolumn{4}{c}{\textit{Test Models}}\\ 
\toprule
\textbf{Victim Model} &  \textbf{$L_\infty$} & EN-B7 Selim & EN-B7 NLab & XN WM & EN-B3 WM\\ 
\midrule
EN-B7 Selim & 0.156 & 100.0\% & 94.5\% & 65.0\% & 68.75\%\\
EN-B7 NLab & 0.156 & 94.5\% & 100.0\% & 75.0\% & 81.50\%\\
XN WM & 0.156 & 77.5\% & 61.0\% & 100.0\% & 20.0\%\\
EN-B3 WM & 0.156 & 66.5\% & 50.5\% & 60.0\% & 100.0\%\\
\bottomrule
\multicolumn{1}{c}{}
\end{tabular}%
}
\caption{Attack success rates of the universal attacks (Section~\ref{sec:universalmethod}) on different victim models and their transferability to unseen detectors (test models).}
\label{tab:universaltransfer}
\end{table}

\subsection{Universal attacks}

To create more accessible attacks, we train a universal adversarial perturbation using the procedure described in Section~\ref{sec:universalmethod}. We set the $L_2$ regularization term $c=0.01$ and use the Adam optimizer with a learning rate of $0.001$. For our initial experiments, we set the the $L_\infty$ threshold $\epsilon=40/255$ for all victim models. 
Since the goal of finding a single input-agnostic perturbation is more challenging than finding one perturbation per video frame, a higher amount of distortion is required for a successful attack as compared to the per-frame attacks described earlier. 
We train the universal perturbation on a dataset of $100$ videos from the DFDC train set which are separate from our evaluation dataset. We train the perturbation using a batch size of $8$ for $10,000$ iterations.

We target one victim model at a time and test the transferability of the universal perturbation on seen and unseen detectors. Table~\ref{tab:universaltransfer} presents the results of performing the universal attack on different victim models at $\epsilon=40/255=0.156$. 
% TODO - add a line on hand tuning.
We are able to achieve 100\% attack  success rate on the same test model as the victim model using a single perturbation across all frames and videos of the same label. 
Also, the universal perturbation is transferable to a significant extent across different models which poses an extremely practical threat to DeepFake detectors in production. Attacking \textit{EN-B7 NLab} results in the most transferable perturbations where we are able to achieve at least a $75\%$ success rate across all unseen detectors.

\begin{figure}[h]
    \centering
    \includegraphics[width=1.0\columnwidth]{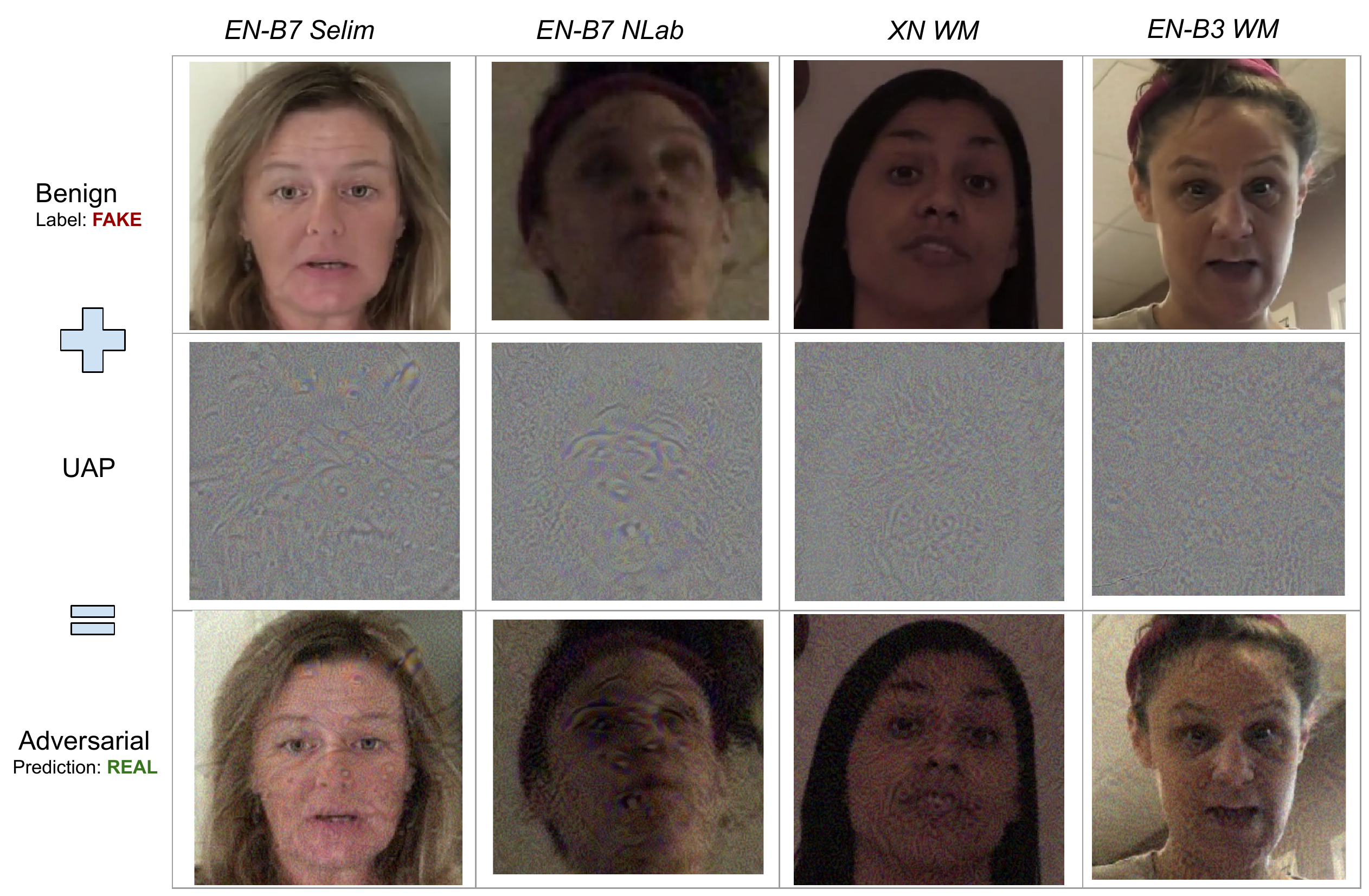}{\centering}
    \caption{Visualization of universal adversarial perturbations trained on different DeepFake detection models.}
    \label{fig:uaps_example}
\end{figure}

Visually, the universal perturbations at $\epsilon=0.156$ are more perceptible than our per-frame attacks discussed in the sections above. Figure~\ref{fig:uaps_example} shows examples of universal adversarial perturbations trained on different DeepFake detectors and the resulting adversarial images obtained after adding the perturbation to the face-crop of the benign frame.

\begin{figure}[b]
    \centering
    \includegraphics[width=1.0\columnwidth]{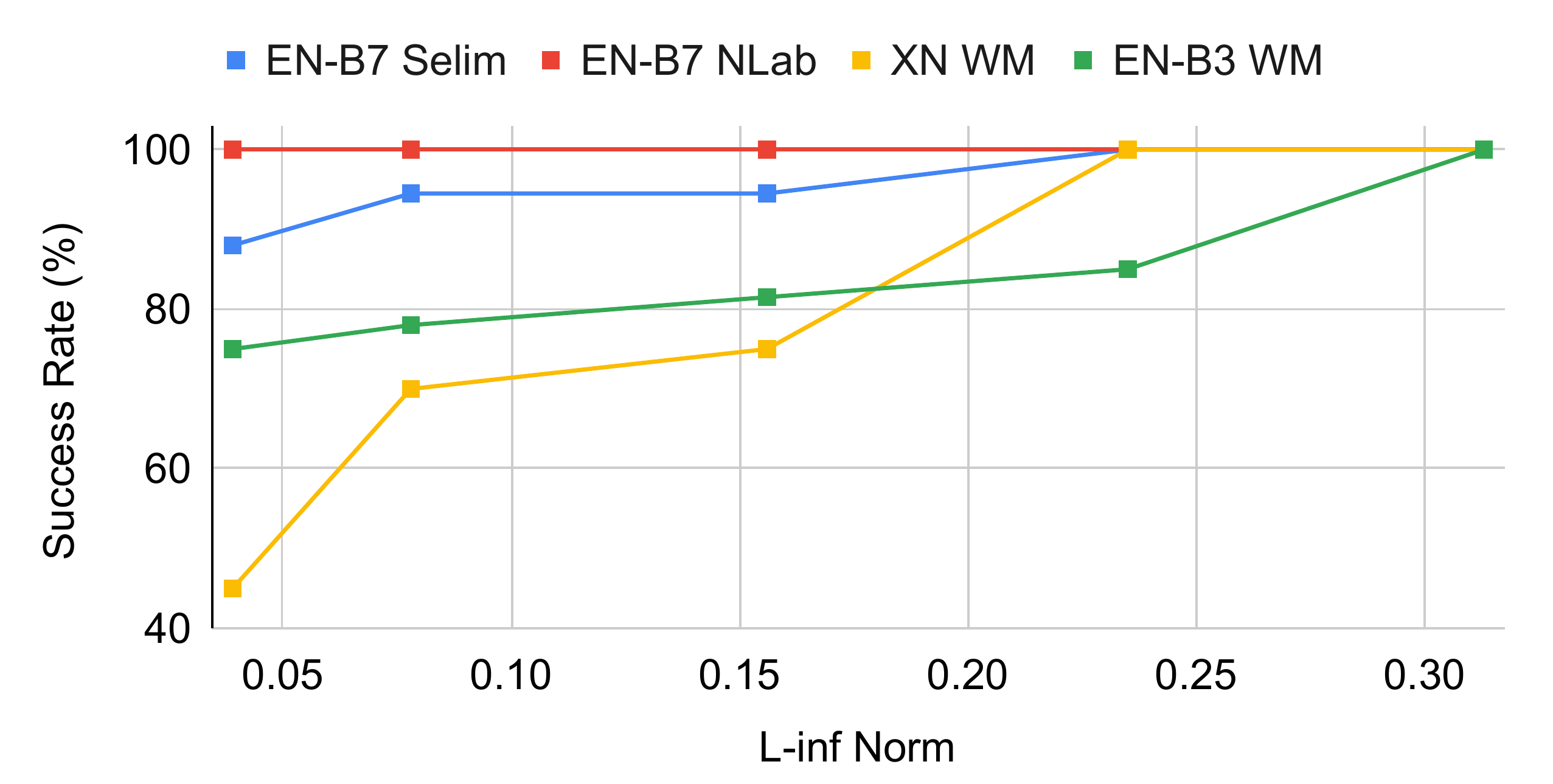}{\centering}
    \caption{Attack success rate on unseen detectors for a universal perturbation trained on the \textit{EN-B7 NLab} detector at different levels of the $L_\infty$ norm of the perturbation.}
    \label{fig:uap_graph}
\end{figure}

\begin{figure}[h]
    \centering
    \includegraphics[width=1.0\columnwidth]{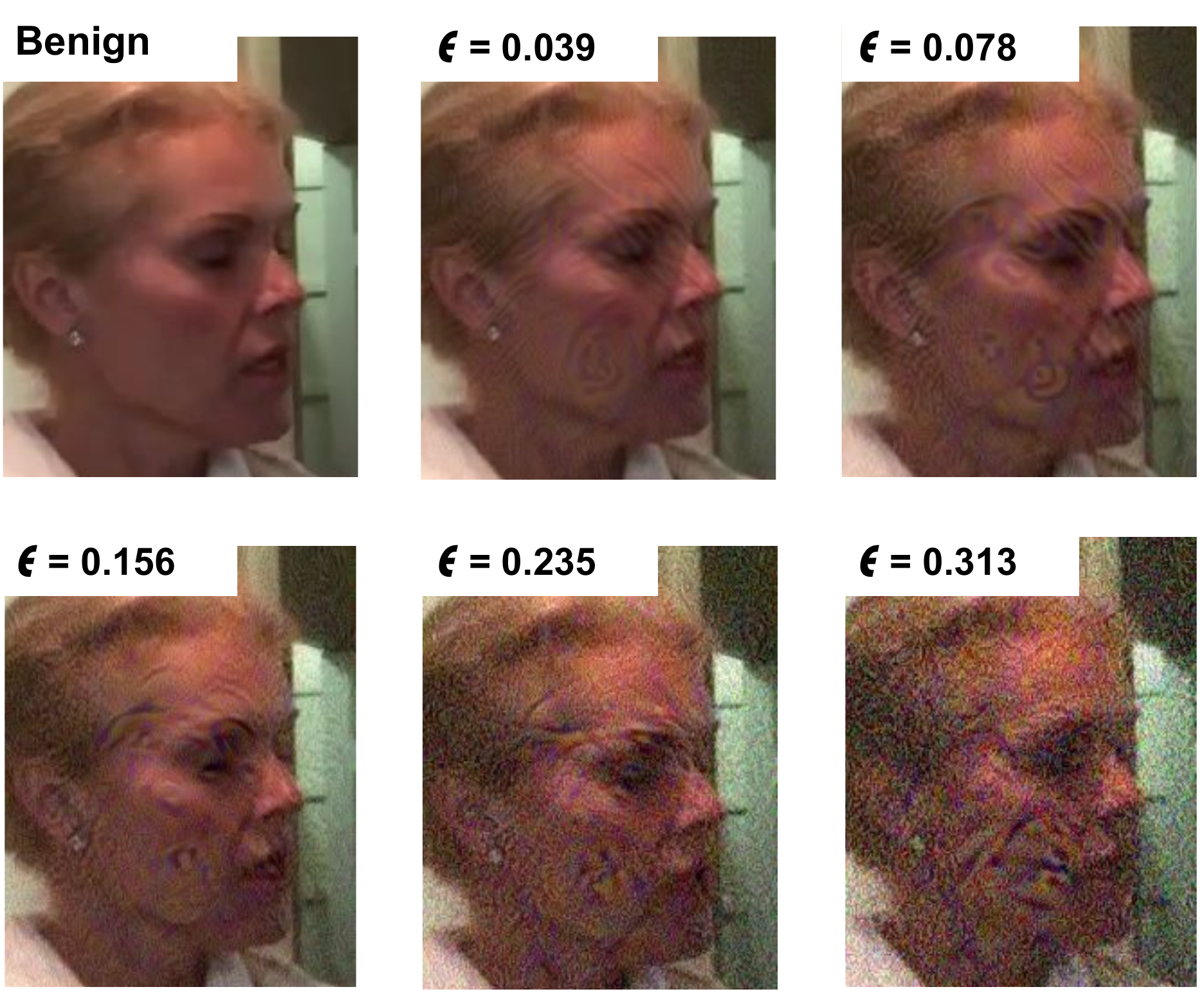}{\centering}
    \caption{Visualization of the perturbed images using different magnitudes ($\epsilon$) of universal adversarial perturbations trained on \textit{EN-B7 NLab}.}
    \label{fig:universalepsilons}
\end{figure}

We perform an additional experiment to study the effectiveness of universal adversarial perturbations at different magnitudes of added perturbations. 
We choose \textit{EN-B7 NLab} as the victim model and perform our universal attack at different values of $\epsilon$. The attack success rates across different models are shown in Figure~\ref{fig:uap_graph}. Figure~\ref{fig:universalepsilons} shows what a perturbed image looks like at different values of $\epsilon$. At $\epsilon < 0.1$, the perturbation is fairly imperceptible but can still achieve high success rates on various test models.

\section{Conclusion}

We design transferable black-box attacks which pose a practical threat to the security of DeepFake detection. 
% We demonstrate that an attacker can bypass state-of-the-art detection systems, without having knowledge of the network parameters and architectures. 
Through our design of universal adversarial perturbations, we demonstrate the ease of accessibility of such attacks since the same perturbation can be added to all the frames of any video in order to bypass multiple detection systems. 
By bypassing recently proposed state-of-the-art DeepFake detection networks with our proposed attack algorithms, this work emphasizes the need for stronger detection methods that are robust to adversaries.

{\small
\bibliographystyle{ieee_fullname}
\bibliography{egbib}
}

\end{document}

%% file: intro.tex
\section{Introduction}
DeepFakes are artificial videos that contain realistically swapped faces mostly created with off-the-shelf neural network based methods. While DeepFakes are sometimes used for humorous or entertainment purposes, these videos are an emerging threat, especially within the realms of politics and misinformation~\cite{emergence}. DeepFakes are especially convincing and have caused harm by making it appear that a prominent person said or did something that they never said or did. Accordingly, effort has been devoted towards training DeepFake classifiers and detectors, which attempt to determine if a video contains a fake face via a variety of methods~\cite{verdoliva2020media}.

\par The state-of-the-art DeepFake detection methods rely on Convolutional Neural Networks (CNNs) to classify a given video as \textit{Real} or \textit{Fake}~\cite{dolhansky2020deepfake}. The best performing methods model the DeepFake detection problem as a per-frame classification problem.
% PAARTH TODO Add a line about the two step detection pipeline
While such methods achieve promising results in terms of detection accuracy, they are vulnerable to adversarial examples~\cite{42503} and can be evaded by adding a carefully crafted perturbation to each frame of a given input video~\cite{carlini2020evading,gandhi2020adversarial, advdeepfakes}. Since DeepFakes have the potential to be 
% so 
very damaging, attacks designed to evade DeepFake detectors can cause outsized harm when compared to other attack scenarios. In addition, as DeepFakes are rare compared to the set of all videos, detection of DeepFakes is already an extremely difficult problem.

\par While adversarial examples pose a threat to DeepFake detectors, designing such examples usually requires complete access to the victim detector model architecture and parameters. 
In a practical threat scenario, the model weights can be kept secret to prevent such white-box attacks. While past works have also proposed black-box attacks to DeepFake video detectors~\cite{advdeepfakes},
they require a large number of queries and access to the model prediction scores for each frame in the video that they aim to misclassify. Such an attack can easily be thwarted by limiting query access and not providing the raw detection scores to the user.  
% This makes the threat infeasible since query access can be limited by the service provider and access to the raw detection scores may not always be available. 

\par Adversarial examples can pose a practical threat to DeepFake detection if they are transferable across different models. Past works have shown that adversarial examples designed to fool a particular network can also fool other networks (with the same or different architecture) trained for the same task~\cite{goodfellow2014explaining,szegedy2013intriguing}. By exploiting this property, an adversary can design attacks on an open source DeepFake detection model and potentially fool a DeepFake detection system in production. 
However, in a real-word scenario, different detection mechanisms employ different input preprocessing steps which can nullify many adversarial attacks. 
% (need citation here). Not sure if past works have talked about this - 
Additionally, DeepFake detectors also use face detection techniques prior to image classification CNNs which can differ across various detection methods. 
This makes it challenging to craft perturbations that are transferable across different detection methods.

% However, in a real-world scenario different detection methods can use different input pre-processing pipelines. 

% In this scenario, an adversary can craft adversarial examples that target a given 

The goal of our work is to study the practical threats posed by adversarial examples to the current state-of-the art DeepFake detection systems. 
To this end, we first study the commonalities between different DeepFake detection methods and gain insight by interpreting the model decisions using gradient-based saliency maps. 
% what the models are looking at while detecting DeepFakes.
We then study the vulnerability of these detection methods to adversarial examples and the extent to which adversarial examples transfer across different detection methods. Next, we propose attack methods that significantly improve the transferability of adversarial examples by designing perturbations that are robust to the differences between the various detection methods. 
% and propose attack methods to improve this trasnferability of adversarial examples thereby posing a signficant practical threat to DeepFake detection. 
Finally, we design more accessible adversarial attacks by creating transferable universal adversarial perturbations that can be universally added across all frames of all videos to reliably fool a number of DeepFake detection methods. 

% Also, designing adversarial examples 

% In this work, we demonstrate several effective attacks against a set of state of the art DeepFake detectors, specifically the top-5 models from the DeepFake Detection Challenge.

% Key strengths:
% 1) Attacks on state of the art video deepfake detectors
% * New challenges
% - Face detection pipeline - different models use different types of face-detectors
% - Input pre-processing
% Robustness to compression important. Different models require different input preprocessing steps.
% - Some models use temporal detection methods

% 2) Black-box accessible evaluation
% - Prior work has either used query based attack for video detectors, 
% or transferibility attack only on image detection - evaluation incomplete. 
% - transferable universal perturbations - new

% 3) Additional explainability section

%% file: cvpr.bbl
\begin{thebibliography}{10}\itemsep=-1pt

\bibitem{afchar2018mesonet}
Darius Afchar, Vincent Nozick, Junichi Yamagishi, and Isao Echizen.
\newblock {MesoNet}: a compact facial video forgery detection network.
\newblock In {\em Workshop on Information Forensics and Security (WIFS)}, 2018.

\bibitem{Amerini_2019_ICCV}
Irene Amerini, Leonardo Galteri, Roberto Caldelli, and Alberto Del~Bimbo.
\newblock Deepfake video detection through optical flow based {CNN}.
\newblock In {\em International Conference on Computer Vision (ICCV)
  Workshops}, 2019.

\bibitem{eot}
Anish Athalye, Logan Engstrom, Andrew Ilyas, and Kevin Kwok.
\newblock Synthesizing robust adversarial examples.
\newblock In {\em International Conference on Machine Learning (ICML)}, 2018.

\bibitem{behjati2019universal}
Melika Behjati, Seyed-Mohsen Moosavi-Dezfooli, Mahdieh~Soleymani Baghshah, and
  Pascal Frossard.
\newblock Universal adversarial attacks on text classifiers.
\newblock In {\em International Conference on Acoustics, Speech and Signal
  Processing (ICASSP)}, pages 7345--7349, 2019.

\bibitem{carlini2020evading}
Nicholas Carlini and Hany Farid.
\newblock Evading deepfake-image detectors with white- and black-box attacks.
\newblock {\em arXiv preprint arXiv:2004.00622}, 2020.

\bibitem{carlini2017towards}
Nicholas Carlini and David Wagner.
\newblock Towards evaluating the robustness of neural networks.
\newblock In {\em Symposium on Security and Privacy}, pages 39--57, 2017.

\bibitem{cheng2019improving}
Shuyu Cheng, Yinpeng Dong, Tianyu Pang, Hang Su, and Jun Zhu.
\newblock Improving black-box adversarial attacks with a transfer-based prior.
\newblock In {\em Advances in Neural Information Processing Systems (NIPS)},
  pages 10934--10944, 2019.

\bibitem{chollet2017xception}
Fran{\c{c}}ois Chollet.
\newblock Xception: Deep learning with depthwise separable convolutions.
\newblock In {\em Conference on Computer Vision and Pattern Recognition
  (CVPR)}, pages 1251--1258, 2017.

\bibitem{nlab}
Azat Davletshin.
\newblock https://github.com/ntech-lab/deepfake-detection-challenge.

\bibitem{retinaface}
Jiankang Deng, Jia Guo, Evangelos Ververas, Irene Kotsia, and Stefanos
  Zafeiriou.
\newblock {RetinaFace}: Single-shot multi-level face localisation in the wild.
\newblock In {\em Conference on Computer Vision and Pattern Recognition
  (CVPR)}.

\bibitem{dolhansky2020deepfake}
Brian Dolhansky, Joanna Bitton, Ben Pflaum, Jikuo Lu, Russ Howes, Menglin Wang,
  and Cristian~Canton Ferrer.
\newblock The {D}eep{F}ake {D}etection {C}hallenge ({DFDC}) dataset.
\newblock {\em arXiv preprint arXiv:2006.07397}, 2020.

\bibitem{dolhansky2019deepfake}
Brian Dolhansky, Russ Howes, Ben Pflaum, Nicole Baram, and Cristian~Canton
  Ferrer.
\newblock The {D}eep{F}ake {D}etection {C}hallenge ({DFDC}) preview dataset.
\newblock {\em arXiv preprint arXiv:1910.08854}, 2019.

\bibitem{gandhi2020adversarial}
Apurva Gandhi and Shomik Jain.
\newblock Adversarial perturbations fool deepfake detectors.
\newblock {\em arXiv preprint arXiv:2003.10596}, 2020.

\bibitem{goodfellow2014explaining}
Ian Goodfellow, Jonathon Shlens, and Christian Szegedy.
\newblock Explaining and harnessing adversarial examples.
\newblock In {\em International Conference on Learning Representations (ICLR)},
  2015.

\bibitem{wm}
Cui Hao.
\newblock https://github.com/cuihaoleo/kaggle-dfdc.

\bibitem{karras2019style}
Tero Karras, Samuli Laine, and Timo Aila.
\newblock A style-based generator architecture for generative adversarial
  networks.
\newblock In {\em Conference on Computer Vision and Pattern Recognition
  (CVPR)}, pages 4401--4410, 2019.

\bibitem{kurakin2016adversarial}
Alexey Kurakin, Ian Goodfellow, and Samy Bengio.
\newblock Adversarial examples in the physical world.
\newblock {\em arXiv preprint arXiv:1607.02533}, 2016.

\bibitem{dsfd}
Jian Li, Yabiao Wang, Changan Wang, Ying Tai, Jianjun Qian, Jian Yang, Chengjie
  Wang, Ji{-}Lin Li, and Feiyue Huang.
\newblock {DSFD:} dual shot face detector.
\newblock In {\em Conference on Computer Vision and Pattern Recognition
  (CVPR)}.

\bibitem{li2019exposing}
Yuezun Li and Siwei Lyu.
\newblock Exposing deepfake videos by detecting face warping artifacts.
\newblock In {\em Conference on Computer Vision and Pattern Recognition (CVPR)
  Workshops}, pages 46--52, 2019.

\bibitem{Liu_2019_ICCV}
Hong Liu, Rongrong Ji, Jie Li, Baochang Zhang, Yue Gao, Yongjian Wu, and Feiyue
  Huang.
\newblock Universal adversarial perturbation via prior driven uncertainty
  approximation.
\newblock In {\em International Conference on Computer Vision (ICCV)}, 2019.

\bibitem{transfer2}
Yanpei Liu, Xinyun Chen, Chang Liu, and Dawn Song.
\newblock Delving into transferable adversarial examples and black-box attacks.
\newblock In {\em International Conference on Learning Representations (ICLR)},
  2017.

\bibitem{universal}
Seyed-Mohsen Moosavi-Dezfooli, Alhussein Fawzi, Omar Fawzi, and Pascal
  Frossard.
\newblock Universal adversarial perturbations.
\newblock In {\em Conference on Computer Vision and Pattern Recognition
  (CVPR)}, 2017.

\bibitem{mopuri2017fast}
Konda~Reddy Mopuri, Utsav Garg, and R~Venkatesh Babu.
\newblock Fast feature fool: A data independent approach to universal
  adversarial perturbations.
\newblock {\em arXiv preprint arXiv:1707.05572}, 2017.

\bibitem{advdeepfakes}
Paarth Neekhara, Shehzeen Hussain, Malhar Jere, Farinaz Koushanfar, and Julian
  McAuley.
\newblock Adversarial deepfakes: Evaluating vulnerability of deepfake detectors
  to adversarial examples.
\newblock {\em arXiv preprint arXiv:2002.12749}, 2020.

\bibitem{neekhara2019universal}
Paarth Neekhara, Shehzeen Hussain, Prakhar Pandey, Shlomo Dubnov, Julian
  McAuley, and Farinaz Koushanfar.
\newblock Universal adversarial perturbations for speech recognition systems.
\newblock In {\em Interspeech}, 2019.

\bibitem{nirkin2019fsgan}
Yuval Nirkin, Yosi Keller, and Tal Hassner.
\newblock {FSGAN}: Subject agnostic face swapping and reenactment.
\newblock In {\em Conference on Computer Vision and Pattern Recognition
  (CVPR)}, pages 7184--7193, 2019.

\bibitem{papernot2016transferability}
Nicolas Papernot, Patrick McDaniel, and Ian Goodfellow.
\newblock Transferability in machine learning: from phenomena to black-box
  attacks using adversarial samples.
\newblock {\em arXiv preprint arXiv:1605.07277}, 2016.

\bibitem{papernot2017practical}
Nicolas Papernot, Patrick McDaniel, Ian Goodfellow, Somesh Jha, Berkay Celik,
  and Ananthram Swami.
\newblock Practical black-box attacks against machine learning.
\newblock In {\em ACM Asia Conference on Computer and Communications Security},
  pages 506--519, 2017.

\bibitem{papernot1}
Nicolas Papernot, Patrick McDaniel, Ian Goodfellow, Somesh Jha, Z~Berkay Celik,
  and Ananthram Swami.
\newblock Practical black-box attacks against machine learning.
\newblock In {\em ACM on Asia Conference on Computer and Communications
  Security}, 2017.

\bibitem{limitations}
Nicolas Papernot, Patrick McDaniel, Somesh Jha, Matt Fredrikson, Z~Berkay
  Celik, and Ananthram Swami.
\newblock The limitations of deep learning in adversarial settings.
\newblock In {\em European Symposium on Security and Privacy (EuroS\&P)}, 2016.

\bibitem{papernot2016distillation}
Nicolas Papernot, Patrick McDaniel, Xi Wu, Somesh Jha, and Ananthram Swami.
\newblock Distillation as a defense to adversarial perturbations against deep
  neural networks.
\newblock In {\em Symposium on Security and Privacy (SP)}, pages 582--597,
  2016.

\bibitem{rahmouni2017distinguishing}
Nicolas Rahmouni, Vincent Nozick, Junichi Yamagishi, and Isao Echizen.
\newblock Distinguishing computer graphics from natural images using
  convolution neural networks.
\newblock In {\em Workshop on Information Forensics and Security (WIFS)}, 2017.

\bibitem{reddy2018ask}
Konda Reddy~Mopuri, Phani Krishna~Uppala, and R Venkatesh~Babu.
\newblock Ask, acquire, and attack: Data-free uap generation using class
  impressions.
\newblock In {\em European Conference on Computer Vision (ECCV)}, pages 19--34,
  2018.

\bibitem{faceforensicsiccv}
Andreas Rossler, Davide Cozzolino, Luisa Verdoliva, Christian Riess, Justus
  Thies, and Matthias Niessner.
\newblock {F}ace{F}orensics++: Learning to detect manipulated facial images.
\newblock In {\em International Conference on Computer Vision (ICCV)}, 2019.

\bibitem{selim}
Selim Seferbekov.
\newblock https://github.com/selimsef/dfdc\_deepfake-\_challenge.

\bibitem{shi2019curls}
Yucheng Shi, Siyu Wang, and Yahong Han.
\newblock Curls \& whey: Boosting black-box adversarial attacks.
\newblock In {\em Conference on Computer Vision and Pattern Recognition
  (CVPR)}, 2019.

\bibitem{tramer}
Dawn Song, Kevin Eykholt, Ivan Evtimov, Earlence Fernandes, Bo Li, Amir
  Rahmati, Florian Tram{\`e}r, Atul Prakash, and Tadayoshi Kohno.
\newblock Physical adversarial examples for object detectors.
\newblock In {\em {USENIX} Workshop on Offensive Technologies}, 2018.

\bibitem{gbb}
J.T. Springenberg, A. Dosovitskiy, T. Brox, and M. Riedmiller.
\newblock Striving for simplicity: The all convolutional net.
\newblock In {\em International Conference on Learning Representations (ICLR)},
  2015.

\bibitem{szegedy2013intriguing}
Christian Szegedy, Wojciech Zaremba, Ilya Sutskever, Joan Bruna, Dumitru Erhan,
  Ian Goodfellow, and Rob Fergus.
\newblock Intriguing properties of neural networks.
\newblock {\em arXiv preprint arXiv:1312.6199}, 2013.

\bibitem{42503}
Christian Szegedy, Wojciech Zaremba, Ilya Sutskever, Joan Bruna, Dumitru Erhan,
  Ian Goodfellow, and Rob Fergus.
\newblock Intriguing properties of neural networks.
\newblock In {\em International Conference on Learning Representations (ICLR)},
  2014.

\bibitem{tan2019efficientnet}
Mingxing Tan and Quoc~V Le.
\newblock Efficientnet: Rethinking model scaling for convolutional neural
  networks.
\newblock {\em arXiv preprint arXiv:1905.11946}, 2019.

\bibitem{verdoliva2020media}
Luisa Verdoliva.
\newblock Media forensics and deepfakes: an overview.
\newblock {\em arXiv preprint arXiv:2001.06564}, 2020.

\bibitem{emergence}
Mika Westerlund.
\newblock The emergence of deepfake technology: A review.
\newblock {\em Technology Innovation Management Review}, 9:40--53, 2019.

\bibitem{zakharov2019few}
Egor Zakharov, Aliaksandra Shysheya, Egor Burkov, and Victor Lempitsky.
\newblock Few-shot adversarial learning of realistic neural talking head
  models.
\newblock In {\em International Conference on Computer Vision (ICCV)}, pages
  9459--9468, 2019.

\bibitem{mtcnn}
K. {Zhang}, Z. {Zhang}, Z. {Li}, and Y. {Qiao}.
\newblock Joint face detection and alignment using multitask cascaded
  convolutional networks.
\newblock {\em IEEE Signal Processing Letters}, 23(10):1499--1503, 2016.

\bibitem{zhou2018transferable}
Wen Zhou, Xin Hou, Yongjun Chen, Mengyun Tang, Xiangqi Huang, Xiang Gan, and
  Yong Yang.
\newblock Transferable adversarial perturbations.
\newblock In {\em European Conference on Computer Vision (ECCV)}, pages
  452--467, 2018.

\end{thebibliography}
